\pdfoutput=1

\documentclass[11pt]{article}

\usepackage[final]{acl}

\usepackage{times}
\usepackage{enumitem}
\usepackage{latexsym}
\usepackage{multirow}
\usepackage[T1]{fontenc}
\usepackage{amsmath}
\usepackage{amsfonts}
\usepackage{float}

\usepackage[utf8]{inputenc}
\usepackage{microtype}
\usepackage{inconsolata}
\usepackage{graphicx}
\usepackage{booktabs}
\usepackage{siunitx,xcolor,rotating,collcell}
\usepackage{colortbl}
\usepackage{amsfonts}
\usepackage{adjustbox}

\definecolor{band}{RGB}{224, 224, 255}

\title{Harnessing Linguistic Dissimilarity for \\
Language Generalization on Unseen Low-Resource Varieties}

\author{
 \textbf{Jinju Kim\textsuperscript{1,2}} 
 \textbf{Haeji Jung\textsuperscript{2,3}} 
 \textbf{Youjeong Roh\textsuperscript{2,4}} 
\\
 \textbf{Jong Hwan Ko\textsuperscript{1}} 
 \textbf{David R. Mortensen\textsuperscript{2}}
\\
 \textsuperscript{1}Department of Electrical and Computer Engineering, Sungkyunkwan University
\\
 \textsuperscript{2}Language Technologies Institute, Carnegie Mellon University
\\
 \textsuperscript{3}Department of Computer Science, University of British Columbia 
 \\
 \textsuperscript{4}Electronics and Telecommunications Research Institute\\
 \small{
   \textbf{Correspondence:} \href{mailto:perla0328@g.skku.edu}{perla0328@g.skku.edu}
   \href{mailto:dmortens@cs.cmu.edu}{dmortens@cs.cmu.edu}
 }
}

\begin{document}
\maketitle
\begin{abstract}
Low-resource language varieties used by specific groups remain neglected in the development of Multilingual Language Models. A great deal of cross-lingual research focuses on inter-lingual language transfer which strives to align allied varieties and minimize differences between them. However, for low-resource varieties, linguistic dissimilarity is also an important cue allowing generalization to unseen varieties. Unlike prior approaches, we propose a two-stage Language Generalization framework that focuses on capturing variety-specific cues while also exploiting rich overlap offered by high-resource source variety. First, we propose TOPPing, a source-selection method specifically designed for low-resource varieties. Second, we suggest a lightweight VAÇAÍ-Bowl architecture that learns variety-specific attributes with one branch while a parallel branch captures variety-invariant attributes using adversarial training. 
We evaluate our framework on structural prediction tasks, which are among the few tasks available, as proxy for performance on other downstream tasks. 
Using VAÇAÍ-Bowl with TOPPing yields an average 54.62\% improvement in the dependency parsing task, which serves as a proxy for performance on other downstream tasks across 10 low-resource varieties.\footnote{Work done while the authors Jinju, Haeji and Youjeong were visiting Carnegie Mellon University.}
\end{abstract}

\section{Introduction}\label{sec:intro}

\begin{figure}
    \centering
    \includegraphics[width=1.01\columnwidth]{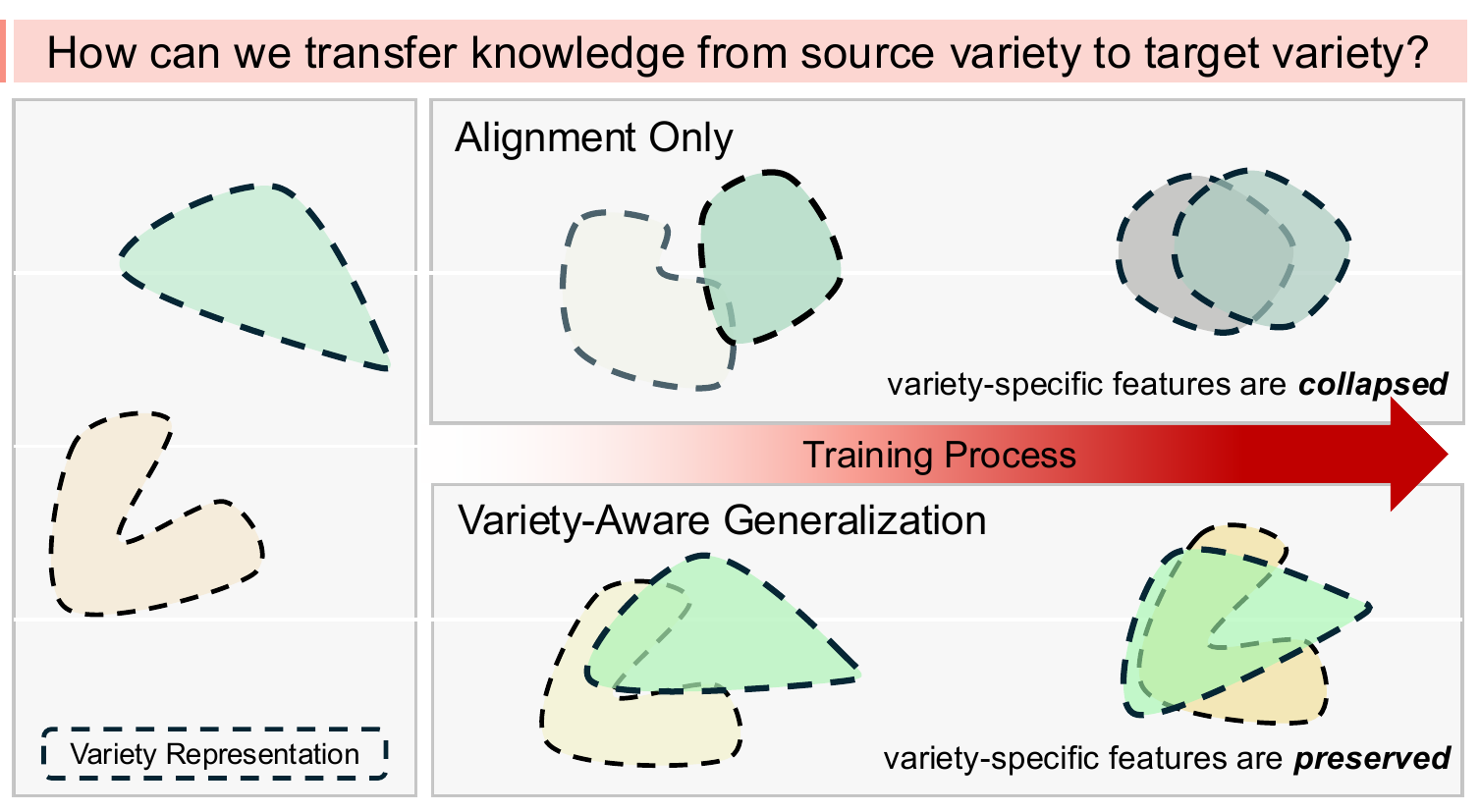}
    \caption{Visualization of training in the embedding space, comparing alignment-based only method and variety-aware generalization method.}
    \label{fig:fig1}
\vskip -0.1in
\end{figure}

Cross-lingual transfer methods overwhelmingly assume that aligning representations across languages is universally beneficial. We show it is not. Multilingual Language Models (MLMs) have extended language technologies to more than one hundred languages \citep{pires-etal-2019-multilingual, conneau-etal-2020-unsupervised}, yet excelling in a language offers little meaning if a model cannot generalize to its variants (e.g., regional dialects). Crucially, a large portion of real-world communication is carried out in language varieties not represented among the roughly one hundred languages that dominate MLM training data. Decades of research in linguistics show that languages lay on a continuum of similarity, each with its own continuous domain \citep{lin-etal-2019-choosing}, and when confronted with such intra-linguistic variants, existing models deliberately fail \citep{faisal-etal-2024-dialectbench}. We find that alignment-only methods exacerbate this problem by collapsing variety-specific structure that models need to generalize (Figure~\ref{fig:fig1}).

Intra-linguistic variation is at least as pervasive as inter-linguistic variation. 
Linguists often refer to intra-language variants as ``dialects'' or ``sociolects''. In this paper, we avoid the term \textit{dialect} for two reasons: (i) it can carry pejorative connotations, and (ii) it is unduly restrictive, implying mutual intelligibility with other variants of a language. We therefore adopt the more neutral and inclusive term \textit{language variety} (or \textit{variety}), that broadly denotes variants shaped by regional, social, and cultural distinctions of its speaker community \citep{chambers1998dialectology}.

Prior research aiming to extend NLP to low-resource varieties, including interlingual transfer, has primarily centered on exploiting cross-variety similarities to align representations (e.g., Figure \ref{fig:fig1}'s variety-aligned only method) to facilitate knowledge transfer \citep{yang2022enhancing}.
However, alignment methods risk losing variety-specific information by collapsing distinct features. This overlooks the linguistic variations that naturally arise in real-world contexts.
In this paper, instead, we propose a variety-aware Language Generalization framework that performs generalization without any training on the target low-resource variety.
By learning not just \textit{how it is similar to a high-resource variety}, but \textit{how it is different}, the model learns to disentangle and strategically combine linguistic features to perform in unseen settings. We also propose an improved automatic method for identifying high-resource varieties most relevant for training a model targeting a particular low-resource variety without any usage of labels, annotation, or parallel dataset. Together, these methods achieve better results than all baselines on structural tasks including dependency parsing (DEP) and part-of-speech tagging (POS), which we believe to be an informative proxy for performance on other downstream tasks.

Our key contributions are as follows:
\begin{itemize}
    \item This paper introduces the Language Generalization framework, focusing on making a model robust to unseen language variations.
    \item We introduce TOPPing, a method for selecting source varieties to generalize on a target low-resource variety without annotations or parallel dataset.
    \item We propose VAÇAI-Bowl, a novel and lightweight architecture to not only align, but also distinguish varieties.
\end{itemize}


\section{Related Work}
\label{sec:related}

\subsection{Low-Resource Varieties}





The disparity of MLMs performing significantly worse in low-resource varieties arises even when the variety is typologically close to, and partially represented in, the training corpus. Through empirical studies, drops in performance when data shifts to a low-resourced variant have been proven to be biased towards dominant varieties \citep{blasi-etal-2022-systematic,blaschke-etal-2024-dialect,blaschke-etal-2023-manipulating,faisal-etal-2024-dialectbench,srivastava-chiang-2025-calling,lin2025languagegapsevaluatingdialect, ustun-etal-2020-udapter}.

Given this limitation, recent approaches leverage closely related high-resource varieties to support lower-resourced counterparts \citep{snaebjarnarson-etal-2023-transfer,bafna-etal-2024-evaluating}. \citet{bafna2025dialupmodelinglanguagecontinuum} train on artificially generated variants or adapt inputs at inference time, while \citet{nguyen2025harnessingtesttimeadaptationnlu} apply test-time adaptation to bridge the dialectal gap.

\subsection{Zero-shot Cross-lingual Transfer}


Prevailing methods to improve cross-lingual transferability without explicit training on the target variety focus on aligning representations via token-level self-augmentation \citep{wang-etal-2023-self-augmentation}, manifold mixup of parallel sentences \citep{yang2022enhancing}, graph-based embedding re-parameterization \citep{wu-monz-2023-beyond}, or robust training with randomized smoothing \citep{huang-etal-2021-improving-zero}. All treat variety-specific information as noise to be suppressed rather than a potential source of transfer knowledge.

Another common strategy is to train models on source languages that are linguistically similar to the target language. 
Prior work investigating the factors that influence transfer performance has shown that linguistic similarity tends to correlate with better cross-lingual transfer \citep{ERONEN2023@talk, ERONEN2023103250,lauscher-etal-2020-zero, dufter-schutze-2020-identifying}. This has motivated efforts to identify optimal source languages using various linguistic similarity metrics. For example, \citet{de-vries-etal-2022-make} examine part-of-speech tagging across diverse source-target language pairs and suggest optimal pairs for certain languages, and
\citet{lin-etal-2019-choosing} proposes a ranking method based on multiple linguistic similarity features, offering a more systematic framework with quantitative features. 
\citet{rice-etal-2025-untangling} utilizes typological and dataset-dependent features to conclude the best rankings for cross-lingual performances.
However, these rely on pre-annotated language information, making it inapplicable for varieties that lack annotated datasets.

Collectively, these methods operate under the implicit assumption that full alignment is universally beneficial. However, this assumption is rarely examined, and existing work provides limited justification for why suppressing variety-specific structure should be the default strategy. Our work directly investigates this gap.





\begin{figure*}[t]
  \includegraphics[width=\textwidth, height=2.8in]{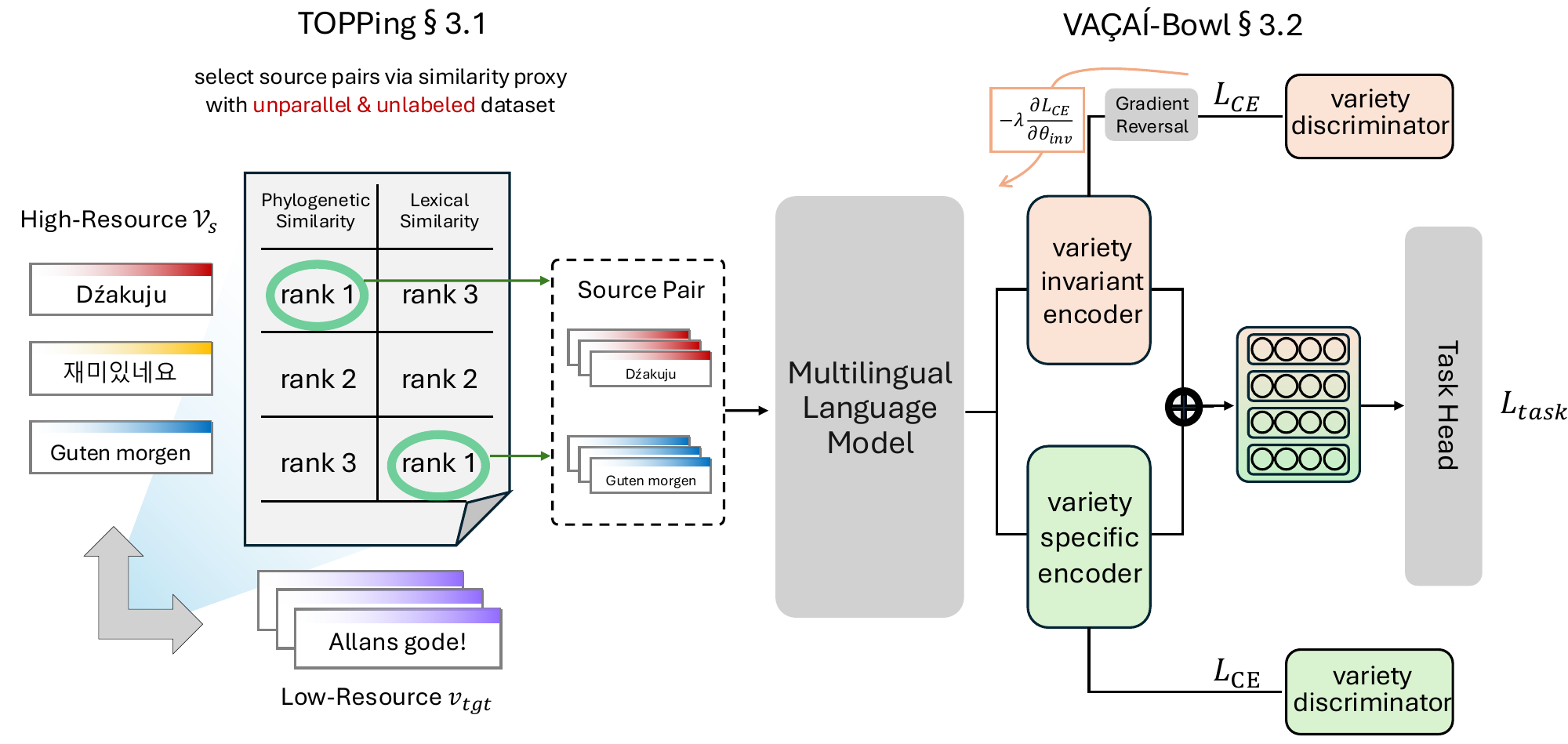}
  \vskip -0.1in
  \caption{Overall framework of this paper. Using TOPPing, with just unparallel and unlabeled datasets, we can select source varieties with not only shared but also distinctive features to capture relationship between varieties. From the obtained source variety pair, VA learns the semantic differences of neighboring source varieties and learns to generalize on the target low-resource variety under unseen setting.}
  \label{fig:framework}
\end{figure*}

\vskip -0.2in

\subsection{Domain Generalization}



Domain Generalization (DG) aims to ensure model performance on domains inaccessible during training \citep{dg2011,dg2013,dg-survey}, typically by learning domain-invariant features \citep{dg2013,dg-adlearning,Li_2018_ECCV}, an approach also adopted in NLP \citep{wang-etal-2024-domain,wang-etal-2021-meta,li-etal-2024-representation}.

In multilingual NLP, prior work treats each language as a separate domain \citep{jung-etal-2024-mitigating}, ensembles adapters for zero-shot generalization \citep{rathore-etal-2023-zgul}, or leverages typological features \citep{adilazuarda-etal-2024-lingualchemy}. Most focus on invariant features \citep{ganin2016domainadversarialtrainingneuralnetworks,ngo2024zeroshotcrosslingualtransferlearning,tahery-etal-2024-cross}; we aim to capture domain-specific signals to help the model recognize linguistic differences within similar varieties---thereby improving generalization to unseen varieties.

\section{Methods}
\label{sec:methods}

In this paper, we propose a novel framework that leverages both variety-invariant and variety-specific information to generalize to unseen low-resource variety, and a method to carefully select exploitable high-resource source variety to pair with the low-resource target variety.

\subsection{TOPPing} \label{sec:methods_1}
For low-resource language varieties, obtaining sufficient labeled training data remains expensive and labor-intensive \cite{blasi-etal-2022-systematic,faisal-etal-2024-dialectbench}.
When the low-resourced target variety has high-resourced neighbors in terms of linguistic similarity, utilizing the latter can be relatively cheaper. 
To mitigate the constraints of low-resource varieties, we introduce \textbf{TOPPing}: \textbf{T}oken-\textbf{O}verlap \& \textbf{P}roximal embedding \textbf{P}air\textbf{ING}, a simple yet effective method that does not require annotations and allows preservation of varietal diversity for selecting source varieties. This allows a wide and general usage even when the variety is unseen, unannotated, and unlabeled. 
Previous works define language distances in various dimensions \citep{littell-etal-2017-uriel, rama-etal-2020-probing}. In this work, we rely on two signals that can be computed automatically from raw text and capture complementary aspects of similarity. As illustrated in Figure \ref{fig:framework}, to encourage diversity in source selection, we compute these similarity signals independently rather than combining them into a single joint score. This preserves the room for variety-specific information training which serves as a key for generalization. 

Let $v_{tgt}$ be the target low-resource variety, and $\mathcal{V}_{\text{src}}$ the set of high-resource source varieties. For each variety $v$, let $\mathbf{X}_v = \{x_i^v\}_{i=1}^{N_v} $ denote its dataset. First, we obtain a source variety by proxying the phylogenetic distance via embedding distance \citep{rama-etal-2020-probing}. Let $f_{\!\text{CLS}_{2}}(x) \in \mathbb{R}^d$ denote the "[CLS]" representation of input $x$ from the second layer of a MLM. We aim to obtain a source variety $v_{\text{sim}} \in \mathcal{V}_{src}$ such that:

\begin{equation}
\begin{split}
v_{\text{sim}} = \underset{v \in \mathcal{V}_{src}}{\arg\min} \;
\left\| \mu_v - \mu_{v_{tgt}} \right\|_2, \\
\mu_v = \frac{1}{N_v} \sum_{x \in \mathbf{X}_v} f_{\text{CLS}_{2}}(x).
\end{split}
\tag{1}
\end{equation}
We use the second-layer "[CLS]" rather than the final layer because lower-layer representations have been shown to capture more typological and morpho-syntactic information, which aligns better to proxy phylogenetic structure \citep{hewitt-manning-2019-structural,mousi-etal-2024-exploring,bakos-etal-2025-alignfreeze}.

Second, we use token overlap \citep{blaschke-etal-2025-add} between two varieties  as a proxy for lexical distance. Here, we introduce \textbf{token-length weighted Jaccard Similarity} as lexical overlap calculation tailored to low-resource varieties. Unlike highly represented varieties, lexical items are often fragmented into shorter sub-tokens, which diminishes the discriminative power of a standard Jaccard similarity measure. Weighting overlaps by token length mitigates this bias, preventing varieties from being erroneously conflated based on scripts. The aim is to obtain a source variety $v_{\text{overlap}} \in \mathcal{V}_{src}$ such that :
\begin{equation}
v_{\text{overlap}} = \underset{v \in \mathcal{V}_{src}}{\arg\max} \;
\operatorname{TJ}(\mathbf{X}_{v}, \mathbf{X}_{v_{tgt}}),
\tag{2}
\end{equation}
where $\operatorname{TJ}(\mathbf{X}_{v}, \mathbf{X}_{v_{tgt}})$ is token-length weighted Jaccard similarity.

\begin{equation}
\begin{split}
\operatorname{TJ}(\mathbf{X}_a, \mathbf{X}_b) = 
\frac{
    \sum_{tok \in T_a \cap T_b} \omega(tok)
}{
    \sum_{tok \in T_a \cup T_b} \omega(tok)
}, \\
\omega(tok) = \max(1, \text{len}(tok) - 1).
\end{split}
\tag{3}
\end{equation}
The pair $\langle v_{\text{sim}}, v_{\text{overlap}}\rangle$, selected by independent ranking, leaves room for diversity in source pair selection. 


\subsection{VAÇAI-Bowl} \label{sec:methods-2}

In Figure~\ref{fig:framework}, we illustrate our approach for Language Generalization : \textbf{V}ariety \textbf{A}ligned and Spe\textbf{C(Ç)}ific \textbf{A}ttr\textbf{I}butes \textbf{B}lending for L\textbf{OW}-resouce \textbf{L}anguage Varieties. This framework leverages both variety-invariant and variety-specific knowledge from high-resource varieties to effectively model representations for an unseen, low-resource variety. 

In order to model variety-invariant and variety-specific features, we use a Multilingual Language Model that produces a "[CLS]" embedding for every input sentence. We implement two independent 2-layer MLP encoders :

\begin{itemize}
    \item Variety-invariant encoder $f_{\text{inv}}$ is trained adversarially to align varieties and learn invariant features. We obtain $h_{\text{inv}}= f_{\text{inv}}([CLS])$.
    \item Variety-specific encoder $f_{\text{spc}}$ is trained normally to emphasize variety-specific features. We can also obtain $ h_{\text{spc}} = f_{\text{spc}}([CLS])$.
\end{itemize}

The outputs are concatenated into $h$, so that $h = h_{\text{inv}} \,\Vert\, h_{\text{spc}} \in \mathbb{R}^{d}$,
where each encoder maps the $d$-dimensional "[CLS]" embedding to a $d/2$-dimensional vector (i.e., 768→384 each), so that the concatenated representation matches the original dimensionality. The joint feature $h$ is used in place of the original \texttt{[CLS]} embedding for downstream tasks.

Our choice of an adversarial objective is motivated by two considerations. First, adversarial alignment is the predominant mechanism for enforcing invariance in both cross-lingual transfer 
\citep{wang-etal-2023-self-augmentation, huang-etal-2021-improving-zero}
and domain generalization. 
\citep{ganin2016domainadversarialtrainingneuralnetworks,ngo2024zeroshotcrosslingualtransferlearning}.
Using the same mechanism allows us to make controlled comparisons with this prior work. Second, our research goal is not to propose a new alignment strategy, but to test whether complementing invariant representations with explicitly modeled variety-specific features improves generalization. The adversarial loss on $f_{\text{inv}}$ ensures that variety-specific signals are excluded from the invariant branch, allowing us to directly evaluate the contribution of the variety-specific encoder $f_{\text{spc}}$.

To train each encoder to extract variety-invariant and variety-specific features, each encoder is paired with its own discriminator ($D_{\text{inv}}$ and $D_{\text{spc}}$) that performs classification on what variety the input belongs to. A gradient-reversal layer $G_{\lambda}$ is inserted in front of $D_{\text{inv}}$ to selectively update parameters to fool $D_{\text{inv}}$ \citep{gradient_rl}.


\begin{equation}
    G_{\lambda}(z) = z, 
    \qquad
    \frac{\partial G_{\lambda}}{\partial \mathbf{z}} = -\lambda\mathbf{I},
    \tag{4}
\end{equation}
where $\lambda$ is a hyperparameter. Contrastingly, $f_{\text{spc}}$ learns to help $D_{\text{spc}}$ by producing easily distinguishable features. The discriminators yield two loss terms : 
\begin{equation}
\begin{aligned}
    L_{\text{inv}} = L_{CE}(D_{\text{inv}}(G_{\lambda}(h_{\text{inv}})), y_{\text{var}}), \\
    L_{\text{spc}} = L_{CE}(D_{\text{spc}}(h_{\text{spc}}), y_{\text{var}}),
\end{aligned}
\tag{5}
\end{equation}
where $L_{\text{CE}}$ denotes Cross Entropy Loss.

Lastly, the task loss is employed for the fine-tuning objective and to ground the representation extraction in right directions. 

\begin{equation}
    L_{\text{task}} = L_{\text{task}}(f_{\text{task}}(h), y_{\text{task}}) .
    \tag{6}
\end{equation}

Finally, the objective function is defined as $L_{\text{total}} = L_{\text{inv}} + L_{\text{spc}} + L_{\text{task}}$.




\section{Experiments}
\label{sec:exp}

In the following section, we provide experiments designed to evaluate the effectiveness of (i) source selection method TOPPing, (ii) framework VAÇAÍ-Bowl. We further investigate (iii) \textit{why linguistic dissimilarities contribute to transfer, and to what extent?} For our experiments, we utilize structured prediction tasks to proxy MLM performance on downstream tasks.

\subsection{Experimental Setup}

\textbf{Benchmark.} DialectBench provides datasets and benchmarks for low-resource varieties with annotations that group varieties into language clusters, allowing direct visualization of performance gaps within the same cluster. For our experiment, we select 10 target low-resource varieties from DialectBench that have no training dataset available for the variety. For source varieties, we utilize the rest of DialectBench and sample high-resourced varieties representative of distinctive language clusters from Universal Dependencies \citep{nivre-etal-2017-universal,de-marneffe-etal-2021-universal}. These source variety sets are used for TOPPing variety selection. We evaluate on dependency parsing (DEP) and part-of-speech tagging (POS) task, which both are structured prediction tasks. We evaluate DEP using Unlabeled Attachment Score (UAS) and Labeled Attachment Score (LAS), which measure correctness of predicted head and head+relation label, respectively. For POS tagging, we report token-level F1. Please refer to Appendix \ref{appx:detailed_exp} for detailed results of downstream tasks across varieties.

\paragraph{Source Selection Baselines.}
In Figure \ref{tab:vacai}, we illustrate how automated source language selection using TOPPing can be a simple yet effective method for source language selection especially for unseen varieties. This selection is applicable to diverse cross-lingual transfer scenarios, not limited to a specific method. As a widely-used baseline, we implement \textbf{LangRank} where the source varieties are selected based on pre-annotated linguistic features and dataset-dependent features \citep{lin-etal-2019-choosing}. 
Please refer to Appendix \ref{appx:selected_langs} for detailed information on selected source varieties for each target variety.

\paragraph{Language Generalization Baselines.}
To compare the VAÇAÍ-Bowl framework, we implement two baselines : (1) \textbf{MLM} baseline illustrates performance when the model is simply finetuned on source languages. (2) \textbf{Alignment} baseline where the model leverages adversarial training to learn only variety-invariant features, adopted from DG and robust training is implemented \citep{huang-etal-2021-improving-zero}.

\begin{table*}[t]
\small
\setlength{\tabcolsep}{4pt} 
  \centering
  \begin{adjustbox}{scale=1, center}
  \begin{tabular}{l|cccccccccc}
  \toprule
\multirow{3}{*}{Methods}&\multicolumn{10}{c}{Varieties}\\ 
\cline{2-11}& aln&  gug&gun& koi&  kpv&lij &nds& sma& gsw&xum  \\ \toprule
\multicolumn{11}{l}{\textit{source is eng}}\\[-0.4em]
\bottomrule
mBERT$^\diamond$& 38.14 & 13.51 & 8.95 & 26.12 & 26.89 & 50.22  &36.77& 19.41 & 36.77 &33.21 \\
mBERT& 39.13 & 13.03  & 12.91 & 30.03 &  29.79 & 49.86 &42.61& 20.81 & 42.49& 32.01 \\ \hline
 \multicolumn{11}{l}{\textit{source selected using LangRank \citep{lin-etal-2019-choosing}}}\\[-0.4em]
 \bottomrule
mBERT&       43.90&        22.13&    10.47&       33.97&        32.51&59.38 &46.45&       27.79&       51.12&36.14 \\
+Alignment&       49.58&     25.98 &  16.40&      36.33 &  33.37 &59.68 & 50.49 &  29.90   &        \underline{52.60} &34.67 \\
+VAÇAÍ-Bowl (OURS) &       \textbf{\underline{51.00}}&  \underline{27.05}    &\underline{17.21}&       \underline{36.90}&       \underline{35.32}&\underline{63.02} &\underline{50.92}&       \underline{32.62}&   52.23  & \underline{36.29} \\  \midrule
 
  \multicolumn{11}{l}{\textit{source selected using TOPPing (OURS)}}\\[-0.4em]
  \bottomrule
mBERT&       44.55& 34.10  &15.18&       40.83&   36.80&63.99 & 
52.54 &       35.63& 57.22   & 37.21 \\
+Alignment&       45.30&     31.56  &16.53&      40.72 & 36.52  &62.96 &52.04&38.80       &     54.84  & 35.99 \\
\rowcolor{band}  +VAÇAÍ-Bowl (OURS)&    
\underline{46.34}&        \textbf{\underline{36.39}} &\textbf{\underline{19.00}}&      \textbf{\underline{42.29}}&   \textbf{\underline{38.19}}     &\textbf{\underline{64.29}} &\textbf{\underline{54.90}}&       \textbf{\underline{39.67}}&   \textbf{\underline{57.74}}   & \textbf{\underline{37.67}} \\ \midrule
\end{tabular}
\end{adjustbox}
  \caption{\label{tab:vacai}
    Quantitative results on UAS scores using mBERT as backbone on dependency parsing task evaluated across low-resource varieties from \citet{faisal-etal-2024-dialectbench}. 
    $^\diamond$ refers to value reported in original paper.
    \underline{Underlined} refers to best performing on target under controlled source variety. \textbf{Bold} refers to best performing on the target variety.
  }
\end{table*}
\begin{table*}[t]
\small
\setlength{\tabcolsep}{4pt} 
  \centering
  \begin{tabular}{l|cccccccccc}
  \toprule
\multirow{3}{*}{Methods}&\multicolumn{10}{c}{Varieties}\\ \cline{2-11}& aln&  gug&gun& koi&  kpv&lij &nds& sma& gsw&xum \\ \toprule
\multicolumn{10}{l}{\textit{source is eng}}\\[-0.4em]
\bottomrule
XLM-R$^\diamond$& 43.50&  11.15& 4.23 & 30.91 & 32.14 & 43.78 &34.70& 28.28& 34.70 &28.75 \\
XLM-R &    52.58&        13.61&4.38&       31.50&        30.60&53.79 &42.92&       30.20&        43.60 &24.50 \\ \hline
 \multicolumn{10}{l}{\textit{source selected using LangRank \citep{lin-etal-2019-choosing}}}\\[-0.4em]
 \bottomrule
XLM-R &      55.58&    28.44    &10.95&       41.51&        33.27&\underline{60.37} &47.36&       41.25&      43.75& 33.08 \\
+Alignment&       57.41&   28.44  &12.47&  40.49 &        35.80 &59.51 &47.61& 42.16&  46.95 & 34.30 \\
+VAÇAÍ-Bowl (OURS)&       \textbf{\underline{58.55}}&  \underline{31.23}    &\underline{12.51}&       \underline{42.41}&  \underline{36.13}    &59.82 &\underline{48.38}&   \underline{42.53}    &      \underline{47.47}  & \underline{34.56}\\ \midrule
 
  \multicolumn{11}{l}{\textit{source selected using TOPPing (OURS)}}\\[-0.4em]
  \bottomrule
XLM-R &  55.07  &   30.57   & 11.27 &       40.50&  \textbf{\underline{38.04}}     &\textbf{\underline{63.80}} &48.73&   40.76 &    52.23 & 35.68\\
+Alignment&  56.54   &  28.67   & 8.47&       39.60&   35.85  &63.13 &50.81&       40.20 &  56.62  & 34.76 \\
\rowcolor{band} +VAÇAÍ-Bowl (OURS)&   \underline{57.50}&  \textbf{\underline{31.97}}  & \textbf{\underline{13.98}} &       \textbf{\underline{44.66}}&   37.76   & 63.44 &\textbf{\underline{51.65}}&    \underline{40.99}   &  \textbf{\underline{57.74}}      &\textbf{\underline{36.60}} \\ \midrule
\end{tabular}
  \caption{\label{tab:vacai-xlm-r}
    Quantitative results on UAS scores using XLM-R as backbone on dependency parsing task evaluated across low-resource varieties from \citet{faisal-etal-2024-dialectbench}.  $^\diamond$ refers to value reported in original paper. \underline{Underlined} refers to best performing on target under controlled source variety. \textbf{Bold} refers to best performing on the target variety.
  }
\end{table*}

\paragraph{Implementation Details.}
We utilize mBERT \citep{bert} and XLM-R \citep{conneau-etal-2020-unsupervised} as MLMs for all tasks. We use the two models in their base size, where mBERT has 110M and XLM-R has 125M number of parameters. Specifically, we utilize publicly available models---\texttt{bert-base-multilingual-cased} for mBERT and \texttt{xlm-roberta-base} for XLM-R---downloaded from Huggingface.\footnote{https://huggingface.co/models} 
We set the learning rate as 2e-4, batch size as 64 for mBERT. We set the learning rate as 5e-5, batch size as 64 for XLM-R. Overall, $\lambda$ for gradient-reversal layer is searched in [0.1, 0.5, 1.0] for the following experiments. Parameters are optimized with Adam optimizer \citep{kingma2015adam}. We finetune each model for 10 epochs and halt at step size of 1000 to not exceed the finetuning steps of zero-shot cross-lingual steps. In accordance with the benchmark evaluation procedure, results are reported from the checkpoint obtaining the highest UAS for the DEP task. Please refer to Appendix \ref{appx:lambda} for detailed information on parameter search regarding lambda.

\subsection{Quantitative Results}

In accordance with the previous discussions, a model that can learn both the variety-invariant and variety-specific features should show higher generalization performance regardless of source varieties. Also, this performance should be boosted with model-agnostic source selection TOPPing that preserves noticeable differences in source varieties. At the same time, TOPPing is also expected to perform comparatively to LangRank which prioritizes linguistic similarities. For scores across LAS evaluation metric, please refer to Appendix \ref{appx:LAS}.

\paragraph{Source Selection.}
Table \ref{tab:vacai} and Table \ref{tab:vacai-xlm-r} presents results on DEP task, using mBERT and XLM-R as backbones, respectively. Comparing the scores reported using each LangRank and TOPPing, it is noticeable that in Table \ref{tab:vacai}, evaluations made using TOPPing outperforms LangRank across all methods in 9 out of 10 varieties. Also, simply finetuning the mBERT model on TOPPing itself beats the best score obtained using LangRank for 8 out of 10 varieties. 
In Table \ref{tab:src_summarized}, this phenomenon is more apparent across additional evaluation metrics and task. TOPPing achieves higher performance enhancement for low-resource varieties overall, for both mBERT and XLM-R.

In cases where LangRank enhances transferability and generalization, Gheg Albanian (aln) and Umbrian (xum),
it is notable that the target varieties all fall into Indo-European family. 

\begin{figure*}[h]
    \centering
    \includegraphics[width=1\textwidth]{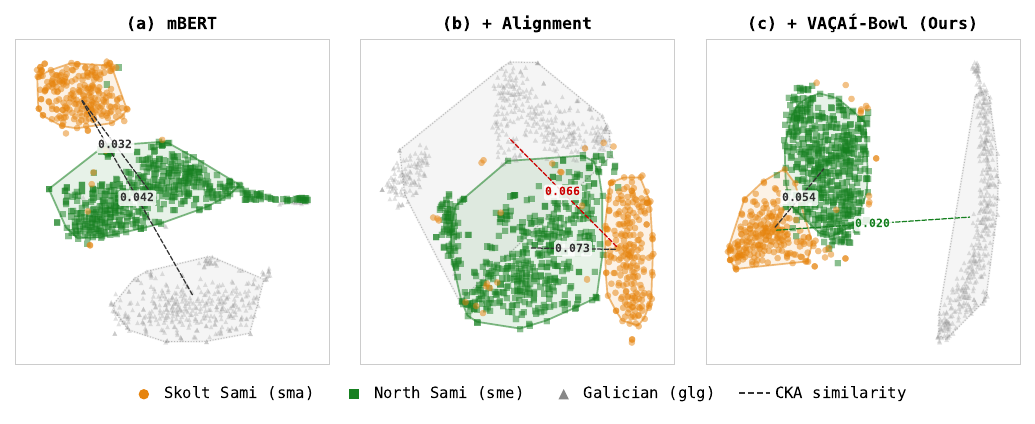}
    \vskip -0.15in
    \caption{t-SNE visualization of sentence embeddings for Skolt Sami (sma), North Sami (sme), and Galician (glg) from final representations. (a)~Pre-trained mBERT separates all three varieties with low inter-variety similarity. (b)~Alignment indiscriminately collapses all representations, including the unrelated Galician. (c)~VA\c{C}A\'{I}-Bowl selectively increases similarity between related varieties while pushing the unrelated Galician further apart.}
    \label{fig:tsne-comparison}
\end{figure*}

\begin{table}[t]
\small
    \centering
    \begin{tabular}{lccc}
        \toprule
         Task & Metric & LangRank & TOPPing  \\
         \toprule
         \multirow{2}{*}{DEP}& UAS & 7.48 & \textbf{10.27} \\
         & LAS & 7.82 & \textbf{10.01} \\
         \midrule
         POS & F1 & 7.36& \textbf{9.34} \\
         \bottomrule
    \end{tabular}
    \caption{
    Comparative results of source selection methods for each task. Each value denotes the \textit{average of absolute improvement in score (in points)} for both mBERT and XLM-R across all methods, compared to the English fine-tuned baseline.}
    \label{tab:src_summarized}
    \vskip -0.1in
\end{table}

\begin{table}[t]
\small
    \centering
    \begin{tabular}{lcccc}
        \toprule
         Task & Metric & MLM & Alignment & VAÇAÍ-Bowl  \\
         \toprule
         \multirow{2}{*}{DEP}& UAS & 7.84 & 8.58 & \textbf{10.21} \\
         & LAS & 8.50 & 8.67 & \textbf{9.58}\\
         \midrule
         POS & F1 & 8.50 & 7.73 & \textbf{8.83} \\
         \bottomrule
    \end{tabular}
    \caption{Comparisons of methods for each task. Each value denotes the \textit{average of absolute improvement in score (in points)} for both mBERT and XLM-R across all methods, compared to the English fine-tuned baseline.}
    \label{tab:met_summarized}
    
\vskip -0.2in
\end{table}

\paragraph{Language Generalization.}
Table \ref{tab:met_summarized} illustrates that our proposed architecture, VAÇAÍ-Bowl, improved model performance regardless of source selection across different evaluation schemes. Specifically, in Table \ref{tab:vacai} VAÇAÍ-Bowl achieves the highest performance on 9 out of 10 target varieties across both source selection methods for mBERT. 
Paying close attention to other baselines, it is notable that the Alignment method, which attempts to enforce alignment by pulling diverse variety embeddings together, fails to surpass the performance on fine-tuned MLM baseline for certain varieties. Specifically, for varieties \{\textit{gug, koi, kpv, lij, nds, gsw, xum}\} in Table \ref{tab:vacai} and \{\textit{gug, gun, koi, kpv, lij, sma}\} in Table \ref{tab:vacai-xlm-r}. 
We refer to this phenomenon as \textbf{alignment-induced failures}. When this occurs, VAÇAÍ-Bowl overcomes the failures of alignment by utilizing variety-specific attributes in all cases. Especially in Table \ref{tab:vacai}, for 6 out of 7 alignment-induced fails observed using TOPPing, VAÇAÍ-Bowl outperforms all methods on the target variety. 
Also, VAÇAÍ-Bowl performs consistently better than Alignment method across all varieties. 

Overall, results presented across low-resource varieties across backbones, tasks, and source selection methods show promising results of VAÇAÍ-Bowl. It can improve model performance up to 27\% (the case of \textit{gug} with LangRank as source selection) of simply finetuning the MLM. We additionally conduct ablation experiments on each component of the loss function, confirming that both 
$\mathcal{L}_{\text{inv}}$ and $\mathcal{L}_{\text{spc}}$ contribute to generalization performance (Appendix~\ref{app:ablation-loss}).

\begin{figure*}[h]
  \includegraphics[width=1.01\textwidth]{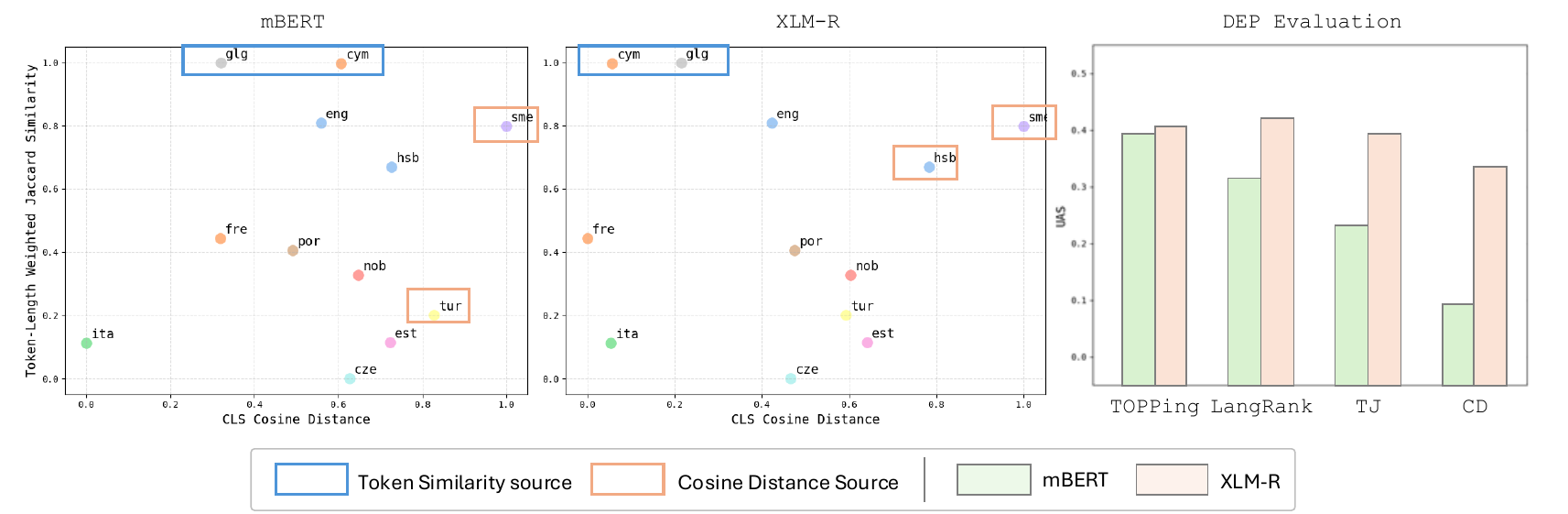}
  \caption{Analysis on source selection method. Two plots on the left illustrate TOPPing source selection scheme on Skolt Sami (sma). The x-axis is closeness of Cosine Distance of "[CLS]" tokens (CD), and the y-axis is token-length weighted Jaccard Similarity (TJ). We experiment VAÇAÍ-Bowl on two more source selections in rightmost plot; Taking two sources with highest TJ and two sources with highest CD. Note that TOPPing selects North Sami (sme), Galician (glg) and LangRank selects Estonian (est), North Sami (sme).}
  \label{fig:sources}
\end{figure*}

\subsection{Qualitative Results}

To examine how each training objective reshapes the representation space, we compute pairwise CKA similarity \cite{kornblith2019similarity} between sentence embeddings of three varieties: Skolt Sami (sma) and North Sami (sme) (related Uralic varieties), and Galician (an unrelated Romance variety). Higher CKA indicates greater representational similarity. Figure~\ref{fig:tsne-comparison} visualizes scores for pre-trained mBERT, the Alignment baseline, and VA\c{C}A\'{I}-Bowl on corresponding embedding spaces via t-SNE.

Alignment uniformly increases CKA across all pairs, including the unrelated Galician, whose similarity to Skolt Sami rises from 0.042 to 0.066. This confirms that alignment collapses representations indiscriminately, regardless of linguistic relatedness.

VA\c{C}A\'{I}-Bowl behaves selectively. It moderately increases similarity between the related Sami varieties (0.032$\to$0.054), reflecting shared Uralic structure captured by the invariant encoder. Crucially, it simultaneously \emph{decreases} similarity of Skolt Sami (sma) to the unrelated Galician (0.042$\to$0.020), indicating that the variety-specific encoder actively separates linguistically distant varieties. This selective behavior of bringing related varieties closer while pushing unrelated ones apart is precisely the property that alignment-only methods lack, and explains why VA\c{C}A\'{I}-Bowl recovers from alignment-induced failures observed in Tables~\ref{tab:vacai}--\ref{tab:vacai-xlm-r}.

Figure \ref{fig:sources} further analyzes how each similarity metric in TOPPing affects source selection for variety \textit{sma}. LangRank selects sources (\textit{est, sme}) that correlate with phylogenetic similarity, yet TOPPing (\textit{glg, sme}) yields stronger performance---suggesting lexical overlap contributes more than phylogenetic proximity alone. The effect is also model-dependent: XLM-R remains robust across diverse source pairs, likely due to its multi-script pre-training, while mBERT benefits more from careful source selection.


\subsection{Analysis}

\paragraph{Q. When is TOPPing most effective?}

TOPPing is most effective for low-resource or under-documented varieties, where typological or lexical metadata are sparse and pre-annotated descriptions are unavailable.
Experimental results show that it remains robust even without such annotations, outperforming methods like LangRank under these conditions.
In contrast, LangRank occasionally performs better for Indo-European family varieties where dense typological and lexical metadata (e.g., from URIEL and WALS) provide rich feature vectors and reliable genealogical signals for ranking candidate sources.
However, for under-documented varieties, every candidate appears equally (dis)similar to LangRank, leading to poor source selection.
Therefore, TOPPing serves as an effective source selection for varieties with minimal annotation and sparse resources.

\paragraph{Q. When is VAÇAÍ-Bowl most effective?}

VAÇAÍ-Bowl is most effective for low-resource varieties, where alignment-based methods tend to fail. For instance, with mBERT backbone (see Table \ref{tab:vacai}), for Umbrian (xum), alignment alone led to a a performance drop in UAS of $36.14$ to $34.67 (-1.47)$ under LangRank and from $37.21$ to $35.99 (-1.22)$ under TOPPing when alignment was applied alone. 
In contrast, incorporating variety-specific features through VAÇAÍ-Bowl raised performance to $36.29 (+1.62)$ and $37.67 (+1.77)$, respectively.
The framework enables the model to overcome alignment-induced failures, showing that enforcing uniform representations across distinct linguistic domains is insufficient. Instead, effective generalization and zero-training transfer require preserving the diversity inherent in each variety.

Overall, contrary to prevailing assumptions in NLP, certain cross-lingual transfer scenarios benefit more from dissimilar language pairs than from closely related ones. These findings underscore the importance of preserving variety-specific information so that models can better generalize to unseen and low-resource varieties, with potential applicability beyond reported cases.

\section{Conclusion}
\label{sec:conclusion}
This paper introduces a Language Generalization pipeline that tackles the twin challenges of selecting helpful high-resource varieties and learning representations that preserve, rather than erase, distinctiveness of varieties. 
Our source-selection strategy, TOPPing, is designed to select two neighboring varieties based on distinct criteria---(i) lexical overlap and (ii) proxy for phylogenetic similarity (using pre-trained MLM). Unlike prior works that utilize pre-annotated information for each variety, our method does not require prior knowledge regarding the target data, ensuring scalability.
Coupled with the lightweight VAÇAÍ-Bowl dual-encoder, one branch aligning varieties and the other amplifying variety-specific cues, our framework delivers consistent gains on dependency parsing. Experiments across ten low-resource varieties, TOPPing with VAÇAÍ-Bowl lifts zero-shot UAS by an average of 50.63\% and 58.6\%, using mBERT and XLM-R, respectively. This beats alignment-centric baseline and even rescues cases where full alignment hurts (“alignment-induced fails”). Beyond parsing, the approach is model-agnostic, computation-friendly, and immediately applicable to other tasks.

\section*{Limitations}
\label{sec:limitations}

The methods presented in this research proved to be effective in handling under-represented varieties that pre-trained MLMs cannot easily generalize to. Although we suggest an end-to-end pipeline that does not require any human annotated work on either source or target varieties, the method still requires unique selection of source varieties for each training. To counterpart this computational complexity, our method trains with only addition of MLP encoders, discriminators, and the task-specific head. This approach significantly reduces both the model size and training overhead compared to methods that require full fine-tuning of MLMs. Yet, it should be recognized that the ultimate goal of Language Generalization is to leverage only a limited set of language varieties to develop a model capable of robust generalization across all varieties, regardless of their resource availability.

\section*{Ethical Considerations}
\label{sec:ethical}

We study a method to enhance zero-shot cross-lingual transfer to very low-resourced varieties, which are often not provided with sufficient data for training or evaluation. While we aim to develop language technologies targeting under-represented language communities, we still lack such coverage, limiting our research to datasets that are publicly available. Nevertheless, our approach provides a valuable step toward addressing the gap, by not simply aligning low-resource varieties with high-resource ones, but instead encouraging the model to recognize and preserve the linguistic differences that define them. 
All resources used in this research are publicly available, and no personal or sensitive information was collected or utilized. We do not anticipate any potential harm arising from this study.

\section*{Acknowledgments}
This work was supported by the Institute of Information \& Communications Technology Planning \& Evaluation (IITP) under multiple grants funded by the Korea government (MSIT), including the ICT Creative Consilience Program (IITP-2026-RS-2020-II201821),
ITRC(Information Technology Research Center)(RS-2021-II212052), and AI Graduate School Support Program (Sungkyunkwan University)(RS-2019-II190421).


\bibliography{anthology, custom}

\onecolumn
\appendix
\section*{Appendix}

\section{Language Codes} \label{appx:language_codes}
In this section, we provide the ISO 639-3 and Universal Dependency dataset code for the varieties used in this paper. 
\begin{table*}[h]
\small
    \centering
    \begin{tabular}{l|c|r}
    \toprule
        Variety & ISO 639-3 & UD-code\\
        \midrule
         gheg & aln & UD\_Gheg-GPS \\
         paraguay mbya guarani & gug & UD\_Mbya\_Guarani-Thomas \\
         brazil mbya guarani & gun & UD\_Mbya\_Guarani-Dooley \\
         permyak komi & koi & UD\_Komi\_Permyak-UH \\
         zyrian komi & kpv & UD\_Komi\_Zyrian-IKDP \\
         ligurian & lij & UD\_Ligurian-GLT \\
         central alemanic & gsw & UD\_Swiss\_German-UZH \\
         skolt saami & sma & UD\_Skolt\_Sami-Giellagas \\
         low saxon & nds & UD\_Low\_Saxon-LSDC \\
         umbrian & xum & UD\_Umbrian-IKUVINA \\
         \bottomrule
    \end{tabular}
    \caption{Target Varieties in Section \ref{sec:exp} with their ISO 639-3 and Universal Dependencies code}
    \label{tab:my_label}
\end{table*}

\begin{table*}[h]
    \small
    \centering
    \begin{tabular}{l|c|r}
    \toprule
        Variety & ISO 639-3 & UD-code\\
        \midrule
         italian & ita & UD\_Italian-MarkIT \\
         norwegian & nor & UD\_Norwegian-Bokmaal \\
         north sami & sme & UD\_North\_Sami-Giella \\
         portuguese & por & UD\_Portuguese-Bosque \\
         spanish & spa & UD\_Spanish-AnCora \\
         finnish & fin & UD\_Finnish-TDT \\
         estonian & est & UD\_Estonian-EDT \\
         catalan & cat & UD\_Catalan-AnCora \\
         indonesian & ind & UD\_Indonesian-CSUI \\
         galician&glg& UD\_Galician-CTG \\
         galician&glg& UD\_Galician-TreeGal \\
         turkish&tur& UD\_Turkish-Penn \\
         turkish&tur& UD\_Turkish-IMST \\
         serbian&srp& UD\_Serbian-SET \\
         croatian&hrv& UD\_Croatian-SET \\
         czech&ces& UD\_Czech-CAC \\
         slovak&slk& UD\_Slovak-SNK \\
         russian&rus& UD\_Russian-SynTagRus \\
         old church slavonic &chu& UD\_Old\_Church\_Slavonic \\
         belarusian&bel& UD\_Belarusian-HSE \\
         ukrainian&ukr& UD\_Ukrainian-IU \\
         upper sorbian&hsb& UD\_Upper\_Sorbian-UFAL \\
         bulgarian&blg& UD\_Bulgarian-BTB \\
         irish&gle& UD\_Irish-IDT \\
         welsh&cym& UD\_Welsh-CCG \\
         \bottomrule
    \end{tabular}
    \caption{Target Varieties in Section \ref{sec:exp} with their ISO 639-3 and Universal Dependencies code}
    \label{tab:my_label_b}
\end{table*}


\newpage

\section{Selected Source Varieties} \label{appx:selected_langs}

In this section, we list the selected source varieties in Section \ref{sec:exp}.
The LangRank selected source varieties are same for both experiments.

\begin{table*}[h]
\small
  \centering
  \begin{tabular}{l|l|c|c|c|c}
  \toprule
    \multirow{2}{*}{cluster} & \multirow{2}{*}{target variety}  & \multicolumn{2}{c|}{LangRank} & \multicolumn{2}{c}{TOPPing} \\
    \cline{3-6}
    & & source & UAS & source & UAS \\ 
    \midrule
        \cline{1-6} \multirow{2}{*}{albanian}&   \multirow{2}{*}{gheg} & italian & \multirow{2}{*}{51.00}& turkish german & \multirow{2}{*}{46.34}\\
        & & finnish& & italian & \\
    \hline
    \multirow{2}{*}{gallo-italian}
        &  \multirow{2}{*}{ligurian} & catalan&  \multirow{2}{*}{63.02}& portuguese & \multirow{2}{*}{64.29}\\
        & & spanish& & italian & \\
    \hline
    \multirow{2}{*}{high german}
        &  \multirow{2}{*}{central alemannic} & norwegian bokmal &  \multirow{2}{*}{52.23}& german & \multirow{2}{*}{57.74}\\
        & & italian & & turkish german & \\
    \hline
    \multirow{4}{*}{komi}
        &  \multirow{2}{*}{komi-zyrian} & indonesian&  \multirow{2}{*}{35.32}& bulgarian & \multirow{2}{*}{38.19}\\
        & & turkish& & russian & \\
        \cline{2-6}
        &  \multirow{2}{*}{komi-permyak} & estonian &  \multirow{2}{*}{36.90}& bulgarian & \multirow{2}{*}{42.29}\\
        & & portuguese& & turkish & \\
    \hline
    \multirow{2}{*}{saami}
        &  \multirow{2}{*}{skolt saami} & estonian&  \multirow{2}{*}{32.62}& north saami & \multirow{2}{*}{39.67}\\
        & & north saami& & galician & \\
    \hline
    \multirow{2}{*}{sabellic}
        &  \multirow{2}{*}{umbrian} & estonian&  \multirow{2}{*}{35.07}& north african arabic & \multirow{2}{*}{37.67}\\
        & & turkish& & italian & \\
    \hline
    \multirow{4}{*}{tupi-guarani} & paraguay & spanish&  \multirow{2}{*}{27.05}& turkish & \multirow{2}{*}{36.39}\\
        & mbya guarani& hindi& & italian & \\
        \cline{2-6}
        &  brazil & finnish&  \multirow{2}{*}{17.21}& english & \multirow{2}{*}{19.00}\\
        & mbya guarani& turkish & & upper sorbian & \\
    \hline
    \multirow{2}{*}{west low german}
        &  \multirow{2}{*}{low saxon} & english&  \multirow{2}{*}{50.92}& turkish german & \multirow{2}{*}{54.90}\\
        & & italian & & english & \\
    
    \bottomrule
\end{tabular}
  \caption{\label{tab:selected_langs}
    Selected two source varieties for Dependency Parsing and its scores with VAÇAÍ-Bowl using mBERT as backbone. Abbreviations (spk) and (wrt) each refer to mode of dataset, spoken and written, respectively.
  }

\vskip -0.2in

\end{table*}
\begin{table*}[h]
\small
  \centering
  \begin{tabular}{l|l|c|c|c|c}
  \toprule
    \multirow{2}{*}{cluster} & \multirow{2}{*}{target variety}  & \multicolumn{2}{c|}{LangRank} & \multicolumn{2}{c}{TOPPing} \\
    \cline{3-6}
    & & source & UAS & source & UAS \\ 
    \midrule
        \cline{1-6} \multirow{2}{*}{albanian}&   \multirow{2}{*}{gheg} & italian & \multirow{2}{*}{58.55}& norwegian bokmaal & \multirow{2}{*}{57.50}\\
        & & finnish& & italian & \\
    \hline
    \multirow{2}{*}{gallo-italian}
        &  \multirow{2}{*}{ligurian} & catalan&  \multirow{2}{*}{59.82}& portuguese & \multirow{2}{*}{63.44}\\
        & & spanish& & italian & \\
    \hline
    \multirow{2}{*}{high german}
        &  \multirow{2}{*}{central alemannic} & norwegian bokmaal &  \multirow{2}{*}{47.47}& german & \multirow{2}{*}{57.74}\\
        & & italian & & turkish german & \\
    \hline
    \multirow{4}{*}{komi}
        &  \multirow{2}{*}{komi-zyrian} & indonesian&  \multirow{2}{*}{35.32}& bulgarian & \multirow{2}{*}{37.76}\\
        & & turkish& & russian & \\
        \cline{2-6}
        &  \multirow{2}{*}{komi-permyak} & estonian &  \multirow{2}{*}{42.41}& bulgarian & \multirow{2}{*}{44.66}\\
        & & portuguese& & turkish & \\
    \hline
    \multirow{2}{*}{saami}
        &  \multirow{2}{*}{skolt saami} & north saami&  \multirow{2}{*}{42.53}& north saami & \multirow{2}{*}{40.99}\\
        & & north saami& & galician & \\
    \hline
    \multirow{2}{*}{sabellic}
        &  \multirow{2}{*}{umbrian} & estonian&  \multirow{2}{*}{34.56}& north african arabic & \multirow{2}{*}{32.77}\\
        & & turkish& & italian & \\
    \hline
    \multirow{4}{*}{tupi-guarani}
        &  paraguay & spanish&  \multirow{2}{*}{31.23}& turkish & \multirow{2}{*}{31.97}\\
        & mbya guarani& hindi& & italian & \\
        \cline{2-6}
        &  brazil & finnish&  \multirow{2}{*}{11.27}& english & \multirow{2}{*}{13.98}\\
        & mbya guarani& turkish & & italian & \\
    \hline
    \multirow{2}{*}{west low german}
        &  \multirow{2}{*}{low saxon} & english&  \multirow{2}{*}{48.38}& german & \multirow{2}{*}{51.65}\\
        & & italian & & turkish german & \\
    
    \bottomrule
\end{tabular}
  \caption{\label{tab:selected_langs-xlm-r}
    Selected two source varieties for Dependency Parsing and its scores with VAÇAÍ-Bowl using XLM-R as backbone. Abbreviations (spk) and (wrt) each refer to mode of dataset, spoken and written, respectively.
  }
\end{table*}

\newpage

\section{Detailed Experimental Results}\label{appx:detailed_exp}

\subsection{Dependency Parsing Results}\label{appx:LAS}

We report UAS scores in Table \ref{tab:vacai} initially for direct comparison to the benchmark DialectBench - that provides only UAS scores. However, to better support the results, we also report the labeled attachment scores (LAS) in this section. 

In Table \ref{tab:vacai-las-mbert} and Table \ref{tab:vacai-las-xlm-r}, LAS scores generally follow the trends of UAS scores reported in the main table. The combination of TOPPing and VAÇAÍ-Bowl enhances the model's ability to capture dependencies in 8 out of 10 low-resource varieties. For the remaining varieties, the following observations hold: For \textit{aln}, VAÇAÍ-Bowl still yields improvements in LAS across both source variety selection methods. For \textit{nds}, the LAS score of TOPPing + VAÇAÍ-Bowl (34.89) remains comparable to the best-performing method (34.96), with UAS scores of 54.90 and 52.54, respectively—indicating a minimal trade-off.

\begin{table*}[h]
\small
\setlength{\tabcolsep}{4pt} 
  \centering
  \begin{adjustbox}{scale=1, center}
  \begin{tabular}{l|cccccccccc}
  \toprule
\multirow{3}{*}{Methods}&\multicolumn{10}{c}{Varieties}\\ 
\cline{2-11}& aln&  gug&gun& koi&  kpv&lij &nds& sma& gsw&xum  \\ \toprule
 \multicolumn{11}{l}{\textit{source selected using LangRank \citep{lin-etal-2019-choosing}}}\\[-0.4em]
 \bottomrule
mBERT&       20.43&        7.54&    1.42&   15.30&  13.17 & 36.89 & 31.44&       13.35&  34.00 & 4.59 \\
+Alignment&    22.00&     7.95 &  1.35&   18.12 &  14.22 & 36.90 & 34.89 &  14.06   &  33.18 & 3.69 \\
+VAÇAÍ-Bowl (OURS) &    22.00&  7.05    & 1.35 &  18.00 &  15.85 &40.21 & 35.25&  15.12&   35.42  & 4.59 \\  \midrule
 
  \multicolumn{11}{l}{\textit{source selected using TOPPing (OURS)}}\\[-0.4em]
  \bottomrule
mBERT&       20.53& 10.41  & 2.32 & 18.11 & 20.33& 42.29 & 
34.96 &   17.19&  37.05   &5.20 \\
+Alignment&    18.21 &   10.74  & 2.21& 18.45 & 20.10  & 43.19 & 32.61 &16.06       & 34.15  & 4.90 \\
\rowcolor{band}  +VAÇAÍ-Bowl (OURS)&    
20.79 &   12.13 &2.44&       18.79&   20.33     &43.79 & 34.89 & 18.21&   37.65   & 5.20 \\ \midrule
\end{tabular}
\end{adjustbox}
  \caption{\label{tab:vacai-las-mbert}
    Quantitative results on LAS scores using mBERT as backbone on dependency parsing task evaluated across selected low-resource varieties from DialectBench.
  }
\end{table*}

\begin{table*}[h]
\small
\setlength{\tabcolsep}{4pt} 
  \centering
  \begin{adjustbox}{scale=1, center}
  \begin{tabular}{l|cccccccccc}
  \toprule
\multirow{3}{*}{Methods}&\multicolumn{10}{c}{Varieties}\\ 
\cline{2-11}& aln&  gug&gun& koi&  kpv&lij &nds& sma& gsw&xum  \\ \toprule
 \multicolumn{11}{l}{\textit{source selected using LangRank \citep{lin-etal-2019-choosing}}}\\[-0.4em]
 \bottomrule
XLM-R&       28.87&  5.34 &1.16& 21.04&  14.80& 38.10 &29.01& 21.12& 27.60 &2.60 \\
+Alignment&    29.73 & 6.23 & 0.81 & 20.02&  16.47& 37.68 &28.98& 19.49& 29.67& 3.52 \\
+VAÇAÍ-Bowl (OURS) &  30.8& 7.46& 3.10& 22.95 &  15.89 &37.43 & 29.37 & 19.53 & 29.39 & 4.75\\  \midrule
 
  \multicolumn{11}{l}{\textit{source selected using TOPPing (OURS)}}\\[-0.4em]
  \bottomrule
mBERT&       27.93&  9.51 &2.38& 20.36&  20.10& 43.29 & 29.82& 18.06 & 35.42 & 5.51 \\
+Alignment&    28.92 &  9.59 &1.08& 21.82 &  20.38 &43.83 & 30.70 & 19.27 & 36.24 & 3.98 \\
\rowcolor{band}  +VAÇAÍ-Bowl (OURS)&    
29.47 & 10.41 &1.10& 22.05&  21.75&43.84 &31.30 & 20.12 & 36.86& 5.21  \\ \midrule
\end{tabular}
\end{adjustbox}
  \caption{\label{tab:vacai-las-xlm-r}
    Quantitative results on LAS scores using XLM-R as backbone on dependency parsing task evaluated across selected low-resource varieties from DialectBench.
  }
\end{table*}

\newpage

\subsection{Part-of-Speech Tagging Results}\label{appx:POS}
In this section, we report the specific evaluated results on POS tagging downstream task for mBERT and XLM-R also represented in Table \ref{tab:met_summarized}. 

\begin{table*}[h]
\small
\setlength{\tabcolsep}{4pt} 
  \centering
  \begin{tabular}{l|cccccccccc}
  \toprule
\multirow{3}{*}{Methods}&\multicolumn{10}{c}{Varieties}\\ \cline{2-11}& aln&  gug&gun& koi&  kpv&lij &nds& sma& gsw&xum \\ \toprule
 \multicolumn{10}{l}{\textit{source selected using LangRank \citep{lin-etal-2019-choosing}}}\\[-0.4em]
 \bottomrule
mBERT &    45.76 & 32.07       & 12.08 & 41.63     &  45.63     & 62.61 & 66.54&  39.93 &  58.64 & 22.61 \\
+Alignment&  47.79 & 32.46    &11.69 &  41.48    & 46.47      & 60.56 & 68.25 & 37.38 & 57.16 &  22.14 \\
+VAÇAÍ-Bowl (OURS)&  45.51 &  32.17 & 14.54 & 42.71 &  45.55 &60.48 & 68.36 & 36.38 & 56.89 &23.81\\
\midrule
 
  \multicolumn{11}{l}{\textit{source selected using TOPPing (OURS)}}\\[-0.4em]
  \bottomrule
mBERT &  43.80 &    33.37   &11.19&  48.68 & 47.71  & 68.32 & 65.48&    42.03 &  59.26 &  26.56 \\
+Alignment&  45.57 & 33.34  & 9.51 &  49.40   & 48.53   & 68.70 & 63.93 &  44.06&  61.22 & 28.26 \\
\rowcolor{band} +VAÇAÍ-Bowl (OURS)& 50.50  & 33.53  & 10.89 & 49.34   & 48.21 & 69.30 & 66.41 & 43.99 &  61.25     &  25.93  \\\midrule
\end{tabular}
  \caption{\label{tab:mbert_pos_f1}
    Quantitative results on F1 scores using mBERT as backbone on part-of-speech tagging task.
  }
\end{table*}

\begin{table*}[h]
\small
\setlength{\tabcolsep}{4pt} 
  \centering
  \begin{tabular}{l|cccccccccc}
  \toprule
\multirow{3}{*}{Methods}&\multicolumn{10}{c}{Varieties}\\ \cline{2-11}& aln&  gug&gun& koi&  kpv&lij &nds& sma& gsw&xum \\ \toprule
 \multicolumn{10}{l}{\textit{source selected using LangRank \citep{lin-etal-2019-choosing}}}\\[-0.4em]
 \bottomrule
XLM-R &56.34&  31.54&5.01& 54.66&  50.87 &62.57 &60.20& 53.83& 48.74&25.28\\
+Alignment& 57.17& 33.37& 6.17 &55.25&  50.48 &61.15 &61.18& 53.11 & 47.40 & 24.64\\
+VAÇAÍ-Bowl (OURS)& 57.97 &  32.94 &6.24& 55.97 &  51.71 &63.54 &60.35& 53.83 & 50.46 & 24.97\\ 
\midrule
 
  \multicolumn{11}{l}{\textit{source selected using TOPPing (OURS)}}\\[-0.4em]
  \bottomrule
XLM-R &56.91&  33.11 & 7.33 & 53.87&  53.73 & 67.98 & 61.95 & 53.89 & 60.16 & \underline{31.07}\\
+Alignment& 57.81 &  34.08 & 4.95 & 55.36 &  53.22 &68.70 & 61.34 & 54.54 & 58.35 & 29.22\\
\rowcolor{band} +VAÇAÍ-Bowl (OURS)& 58.35 &  34.88 & 5.47&  54.02 &  55.16 & 69.97 &60.81& 55.17& 60.21 & 29.23\\
\midrule
\end{tabular}
  \caption{\label{tab:xlm-r-pos-f1}
    Quantitative results on F1 scores using XLM-R as backbone on part-of-speech tagging task. 
  }
\end{table*}


\section{Parameter Search for Lambda of Gradient-Reversal Layer}\label{appx:lambda}

\begin{table*}[h]
    \small
    \centering
    \begin{tabular}{l|cccccccccc}
         \toprule
         lambda & aln & gug & gun & koi & kpv & lij & nds & sma & gsw & xum  \\
         \midrule
         0.1 & 46.09 & 35.00 & 15.03 & 36.56 &40.61&64.72&52.13&37.18&56.40&36.14 \\
         0.5 &45.41 &35.25&15.17&35.70&40.27&63.74&52.36&40.12&54.91&37.37 \\
         1.0 & 46.34 & 36.39 & 19.00 & 42.29 & 38.19 & 64.29 & 54.90 & 39.67 & 57.74 & 37.67 \\
         \bottomrule
    \end{tabular}
    \caption{Ablation study on VAÇAÍ-Bowl performance with mBERT backbone based on different lambda values.}
    \label{tab:lambda}
\end{table*}

For the gradient-reversal layer used to adversarially train invariant feature encoder in Section \ref{sec:methods-2}, we provide an ablation study on its affects on performance.

\newpage

\section{Ablation on Loss Components}
\label{app:ablation-loss}

\begin{table*}[h]
    \centering
    \small
    \begin{tabular}{l ccc | cccccccccc}
        \toprule
         & $\mathcal{L}_{\text{inv}}$ & $\mathcal{L}_{\text{spc}}$ & $\mathcal{L}_{\text{task}}$ 
         & aln & gug & gun & koi & kpv & lij & nds & sma & gsw & xum \\
        \midrule
        w/o both 
            & -- & -- & \checkmark 
            & 45.99 & 34.02 & 15.60 & 37.35 & 36.19 & 62.69 & 51.76 & 37.82 & 54.84 & 37.21 \\
        w/o $\mathcal{L}_{\text{spc}}$ 
            & \checkmark & -- & \checkmark 
            & 46.49 & 29.19 & 15.35 & 40.27 & 36.08 & 63.17 & 53.05 & 35.56 & 56.25 & 26.27 \\
        w/o $\mathcal{L}_{\text{inv}}$ 
            & -- & \checkmark & \checkmark 
            & \textbf{46.74} & 20.82 & 15.44 & 40.33 & 36.95 & 63.23 & 53.20 & 38.24 & 54.69 & 27.18 \\
        \midrule
        \textbf{Full} 
            & \checkmark & \checkmark & \checkmark 
            & 46.34 & \textbf{36.39} & \textbf{19.00} & \textbf{42.29} & \textbf{38.19} & \textbf{64.29} & \textbf{54.90} & \textbf{39.67} & \textbf{57.74} & \textbf{37.67} \\
        \bottomrule
    \end{tabular}
    \caption{Ablation study on loss components of VA\c{C}A\'{I}-Bowl using mBERT with TOPPing source selection (UAS). 
    Removing $\mathcal{L}_{\text{inv}}$ disables adversarial invariance training; 
    removing $\mathcal{L}_{\text{spc}}$ removes the variety-specific objective. 
    The full objective consistently outperforms all ablated variants, 
    confirming that both branches contribute to generalization.}
    \label{tab:ablation-loss}
\end{table*}

To isolate the contribution of each loss term in VA\c{C}A\'{I}-Bowl, we conduct an ablation study by selectively removing $\mathcal{L}_{\text{inv}}$ (adversarial invariance) and $\mathcal{L}_{\text{spc}}$ (variety-specific discrimination) 
from the full objective $\mathcal{L}_{\text{total}} = \mathcal{L}_{\text{inv}} + \mathcal{L}_{\text{spc}} + \mathcal{L}_{\text{task}}$. All experiments use mBERT 
with TOPPing source selection.

Table~\ref{tab:ablation-loss} reports UAS scores across all ten target varieties. Removing $\mathcal{L}_{\text{spc}}$ leads to notable degradation on varieties such as gug and sma, where variety-specific cues are most beneficial. Removing $\mathcal{L}_{\text{inv}}$ similarly degrades performance, indicating that enforcing invariance on one branch is necessary for the other to capture complementary 
variety-specific structure. Training with only $\mathcal{L}_{\text{task}}$ (no disentanglement) yields the weakest average, validating the dual-encoder design. 
The full objective achieves the highest average UAS (43.65), confirming that both branches are essential for effective generalization.

\end{document}